\documentclass[sigconf, screen]{acmart}

\renewcommand\footnotetextcopyrightpermission[1]{} %
\usepackage{graphicx} %
\usepackage{natbib}  %
\usepackage{caption} %
\usepackage{algorithm}
\usepackage{listings}
\usepackage{algorithmic}
\usepackage{amsmath}
\usepackage{booktabs}
\usepackage{multirow}
\usepackage[outdir=./]{epstopdf}
\usepackage{enumitem}
\usepackage{caption}
\usepackage{subcaption}
\usepackage{newfloat}
\usepackage{float}
\usepackage{balance}
\usepackage{svg} %
\usepackage[normalem]{ulem}
\usepackage{threeparttable}
\usepackage{multirow} %
\usepackage{xcolor}
\usepackage{colortbl}   %
\usepackage{xcolor}     %

\definecolor{darkred}{rgb}{0.5,0,0}
\definecolor{darkblue}{rgb}{0,0,0.5}
\definecolor{darkgreen}{rgb}{84,130,53}

\AtBeginDocument{%
  }

\begin{document}

\title{Controllable and Stealthy Shilling Attacks via Dispersive \\Latent Diffusion}

\author{Shutong Qiao}
\affiliation{
  \institution{University of Queensland}
  \city{Brisbane}
  \country{Australia}
}
\email{shutong.qiao@uq.edu.au}

\author{Wei Yuan}
\affiliation{
  \institution{University of Queensland}
  \city{Brisbane}
  \country{Australia}
}
\email{w.yuan@uq.edu.au}

\author{Junliang Yu}
\affiliation{
  \institution{University of Queensland}
  \city{Brisbane}
  \country{Australia}
}
\email{jl.yu@uq.edu.au}

\author{Tong Chen}
\affiliation{
  \institution{University of Queensland}
  \city{Brisbane}
  \country{Australia}
}
\email{tong.chen@uq.edu.au}

\author{Quoc Viet Hung Nguyen}
\affiliation{
    \institution{Griffith University}
    \city{Gold Coast}
    \country{Australia}
}
\email{henry.nguyen@griffith.edu.au}

\author{Hongzhi Yin}
\authornote{Corresponding author.}
\affiliation{
    \institution{University of Queensland}
    \city{Brisbane}
    \country{Australia}
}
\email{h.yin1@uq.edu.au}

\renewcommand{\shortauthors}{Shutong Qiao et al.}
\begin{abstract}

Recommender systems (RSs) are now fundamental to various online platforms, but their dependence on user-contributed data leaves them vulnerable to shilling attacks that can manipulate item rankings by injecting fake users.
Although widely studied, most existing attack models fail to meet two critical objectives simultaneously: achieving strong adversarial promotion of target items while maintaining realistic behavior to evade detection. As a result, the true severity of shilling threats that manage to reconcile the two objectives remains underappreciated.
To expose this overlooked vulnerability, we present DLDA, a diffusion-based attack framework that can generate highly effective yet indistinguishable fake users by enabling fine-grained control over target promotion.
Specifically, DLDA operates in a pre-aligned collaborative embedding space, where it employs a conditional latent diffusion process to iteratively synthesize fake user profiles with precise target item control. To evade detection, DLDA introduces a dispersive regularization mechanism that promotes variability and realism in generated behavioral patterns.
Extensive experiments on three real-world datasets and five popular RS models demonstrate that, compared to prior attacks, DLDA consistently achieves stronger item promotion while remaining harder to detect. These results highlight that modern RSs are more vulnerable than previously recognized, underscoring the urgent need for more robust defenses.
\end{abstract}



\maketitle
\section{Introduction}
Recommender systems (RSs) \cite{resnick1997recommender, burke2011recommender} have become integral to modern online platforms, guiding users toward relevant products, content, and information within vast digital ecosystems. 
To improve recommendation quality, a wide range of techniques have been developed. Among them, Collaborative Filtering (CF) methods, such as matrix factorization and graph neural network-based approaches~\cite{herlocker2000explaining, su2009survey, zhang2016collaborative}, have consistently demonstrated strong performance in both accuracy and efficiency. 
Even within the recent surge of interest in Large Language Model (LLM)-based recommenders, CF-based models continue to serve as the backbone of industrial RSs, particularly in high-throughput stages like pre-ranking over large-scale item pools~\cite{kim2024large}.

Although CF-based systems offer many benefits, their reliance on user-item interaction data makes them vulnerable to shilling attacks~\cite{gunes2014shilling, si2020shilling, wu2012hysad}, where adversaries inject fake user profiles to manipulate recommendations, undermining fairness and trust and degrading user experience. For example, a vendor on an e-commerce platform may promote a less competitive headphone by creating fake users that rate it highly and co-rate popular electronics. These manipulated accounts mislead the recommender into associating the target item with trending products, resulting in its widespread exposure to genuine users regardless of actual relevance.

To prevent such manipulations, numerous efforts have sought to reveal the vulnerabilities of CF-based RSs by crafting various shilling attacks, which are typically categorized into heuristic, opti\-mization-based, and generative approaches.
Heuristic methods, such as Random and Bandwagon attacks, represent some of the earliest attempts at shilling. However, they often fail to produce effective attacks due to their disregard for the collaborative signals embedded in user-item interactions.
Optimization-based methods (e.g., GSPAttack~\cite{nguyen2023poisoning}, CLeaR~\cite{wang2024unveiling}) address this by optimizing fake profiles through model gradients to promote target items. However, they typically lack consideration for stealth, resulting in abnormal behaviors that can be easily detected.
\begin{figure}
    \centering
    \includegraphics[width=0.6\linewidth]{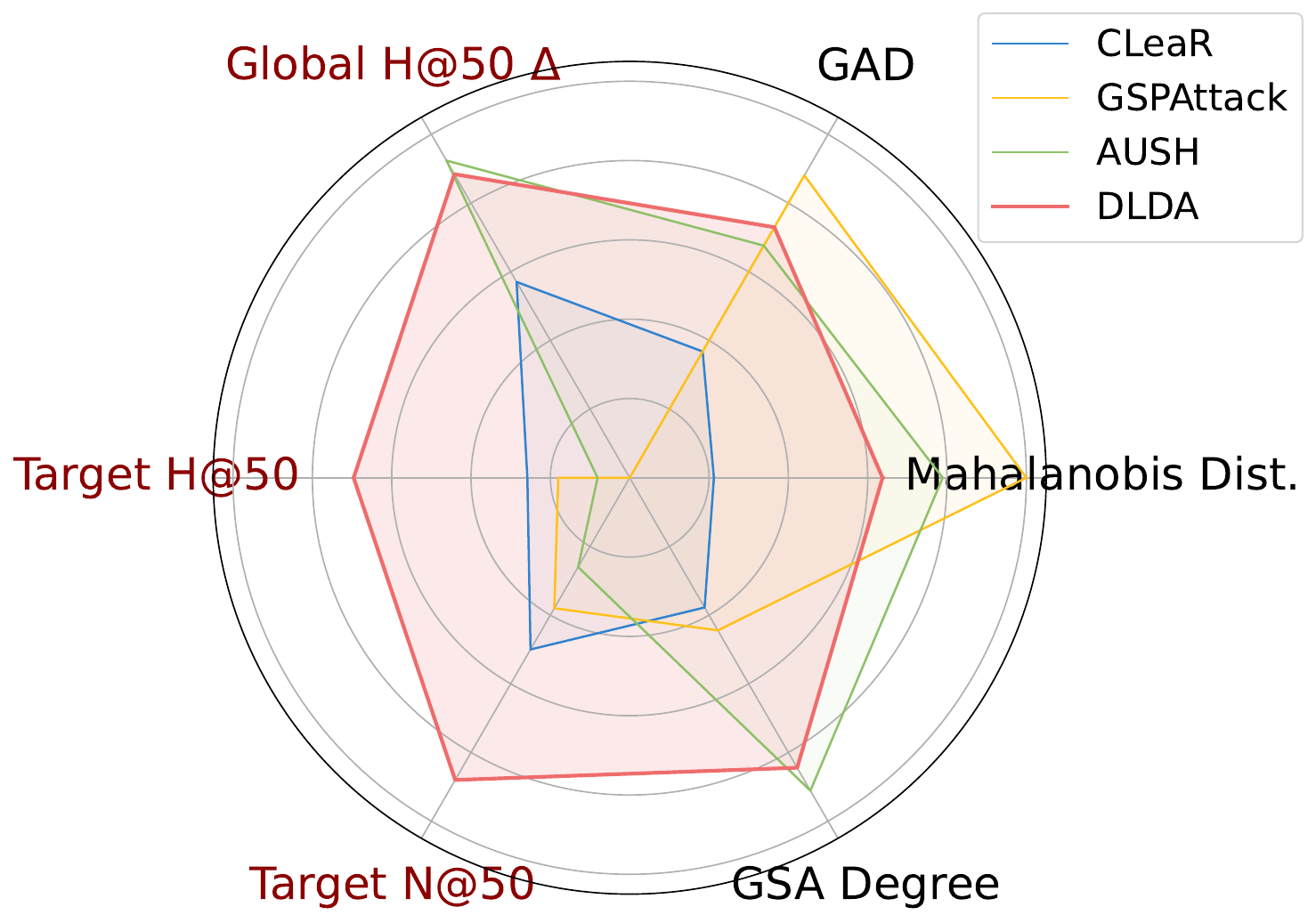}
    \caption{Comparison of attack stealthiness and effectiveness of DLDA and three prior attack models across multiple metrics. Red axis labels indicate effectiveness; black labels indicate stealthiness. See Appendix~\ref{appendix: C} for metric details.}
    \label{fig:intro}
\end{figure}

More recently, generative methods have emerged as the most advanced class of shilling attacks, leveraging generative models (e.g., GAN~\cite{goodfellow2014generative}, VAE~\cite{kingma2013auto}, and diffusion models~\cite {ho2020denoising}) to synthesize user profiles. While promising, these approaches still struggle to reconcile attack effectiveness with stealthiness. 
For instance, AUSH~\cite{lin2020attacking} adopts a GAN-based framework that jointly optimizes shilling and reconstruction objectives. Despite its strong target promotion effect, it often suffers from mode collapse, generating repetitive and unrealistic user profiles that are easily detected. ToDA~\cite{liu2024toda} alleviates these limitations by performing diffusion in the latent space and introducing target item templates to guide generation. This enables controllable target-directed attacks. However, the resulting fake users tend to cluster unnaturally around the target signal, reducing pattern variability and compromising stealth.
As illustrated in Figure~\ref{fig:intro}, existing shilling attacks typically sacrifice effectiveness due to less controllability or stealth, limiting their ability to reveal the true vulnerabilities of CF-based recommenders.

To expose this overlooked vulnerability and improve RSs' robustness, we propose Dispersive Latent Diffusion Attack (DLDA), which is grounded in the key principle that plausible fake profiles should mimic the behavioral patterns embedded in genuine user-item interactions, while still allowing precise control over target item promotion for attack effectiveness. To ensure behavioral realism, DLDA operates within a pre-aligned collaborative latent space, ensuring structural consistency with the underlying CF model. It further incorporates a dispersive regularization term that encourages dispersion across latent representations, preventing over-concentration of the generated users. Additionally, DLDA applies an adaptive Poisson-based projection to convert dense latent vectors into sparse interaction profiles to mimic real-world interaction sparsity.
To enable fine-grained target control, DLDA adopts a conditional latent diffusion process that iteratively refines noise into structured user representations. At each denoising step, a dual cross-attention mechanism injects two complementary signals: the target item embedding guides generation toward the promotional goal, while the user embedding, obtained from a real user who interacted with the target item, encourages behaviors that reflect typical user preferences associated with the target item. Together, these complementary signals enable the model to generate interaction behaviors that are both target-driven and user-consistent.

To sum up, our contributions are as follows:
\begin{itemize}[leftmargin=*]
    \item We propose DLDA, a novel diffusion-based attack framework that synthesizes fake users with both high attack effectiveness and strong stealth under black-box settings.

    \item Unlike prior attacks that trade off realism for control, we show that it is possible to reconcile both through a unified design, uncovering a stronger threat model than previously recognized.

    \item We conduct extensive experiments on multiple benchmarks and victim models, demonstrating that DLDA consistently achieves superior target promotion while evading multiple detection methods, motivating more robust defense strategies for CF.
\end{itemize}

\section{Related Work}
\label{sec:related}
\subsection{Model-based Collaborative Filtering}
Model-based CF remains the dominant paradigm in modern RSs, striking a balance between accuracy, interpretability, and scalability. These methods learn latent representations of users and items from historical interactions to predict future preferences. Early approaches such as Matrix Factorization (MF) \cite{he2016fast} model user-item interactions as inner products in a shared latent space. To capture more complex preference patterns, NCF \cite{he2017neural} incorporates multilayer perceptrons, enabling non-linear modeling. Graph-based models further advance CF by exploiting higher-order connectivity in user-item bipartite graphs. Notably, NGCF \cite{wang2019neural} incorporates high-order connectivity by propagating embeddings on a user-item bipartite graph using graph neural networks (GNNs), but at the cost of added architectural complexity, while LightGCN \cite{he2020lightgcn} simplifies the architecture by retaining only linear neighborhood aggregation, becoming a widely used baseline due to its efficiency and performance.

Recent works improve representation learning in CF via self-supervised contrastive learning. SimGCL\cite{yu2022graph} and XSimGCL~\cite{yu2023xsimgcl} inject random or adaptive perturbations into embeddings to enhance robustness without requiring graph augmentations. Beyond pairwise interactions, HCCF~\cite{xia2022hypergraph} leverages hypergraphs to model multi-way user-item relations, while BIGCF~\cite{zhang2024exploring} and SCCF~\cite{wu2024unifying} unify individuality, collectivity, and contrastive signals in joint frameworks. EGCF~\cite{zhang2024simplify} further pushes simplification by removing explicit embeddings.
To improve semantic modeling and generalization, IPCCF\cite{qin2025polycf} introduces intent-aware contrastive learning with global message propagation for disentanglement. PolyCF\cite{li2025intent} reformulates CF as graph signal recovery using polynomial spectral filters to capture multi-frequency patterns. LightCCF\cite{zhang2025unveiling} bridges contrastive learning and neighborhood aggregation, achieving high efficiency and effectiveness without augmentations or deep GNNs.

Given their wide adoption and impact, enhancing the security of CF-based recommenders against adversarial threats is essential.

\subsection{Targeted Shilling Attack}

\subsubsection{Black-box Attacks}
Black-box poisoning attacks assume no access to the victim recommender’s parameters or gradients.
Early heuristic methods like Random~\cite{lam2004shilling} and Bandwagon~\cite{gunes2014shilling} co-rate target and popular items but lack fine control and are easily detected.
AUSH~\cite{lin2020attacking} enhances stealth by augmenting genuine user profiles, yet fails to align injected behavior with target semantics.
PoisonRec~\cite{song2020poisonrec} adopts a model-free reinforcement learning framework that learns attack policies from reward signals, without relying on victim model knowledge.
GSPAttack~\cite{nguyen2023poisoning} employs a surrogate GNN to optimize fake profiles via gradients, but its success hinges on surrogate-victim similarity and lacks diversity control, limiting transferability.
ToDA~\cite{liu2024toda} introduces a diffusion-based generator for target-oriented attacks under black-box constraints, but it does not explicitly regularize intra-group diversity, making fake users more detectable under distribution shifts.

\subsubsection{White-box Attacks}
White-box attacks assume full access to the model’s architecture and gradients.
PGA \cite{li2016data} formulates a bi-level optimization to poison training data for target item promotion, while FedRecAttack \cite{rong2022fedrecattack} and PipAttack \cite{zhang2022pipattack} perform client-level attacks using local gradient information in federated recommenders.
RAPU \cite{zhang2021data} introduces a novel bi-level optimization framework that incorporates a probabilistic generative model to identify users and items with sufficient interaction signals and minimal perturbation. 
CLeaR \cite{wang2024unveiling} targets contrastive learning-based recommenders, crafting adversarial representations with CW-style losses to maximize target exposure while maintaining global performance.

While white-box methods achieve strong effectiveness, their reliance on privileged access limits real-world applicability. In contrast, existing black-box attacks often trade off target controllability or stealth. Motivated by these limitations, we propose DLDA, which leverages a collaboratively aligned latent diffusion process to generate fake users with precise target control and high stealth.

\begin{figure*}
    \centering
    \includegraphics[width=0.92\textwidth]{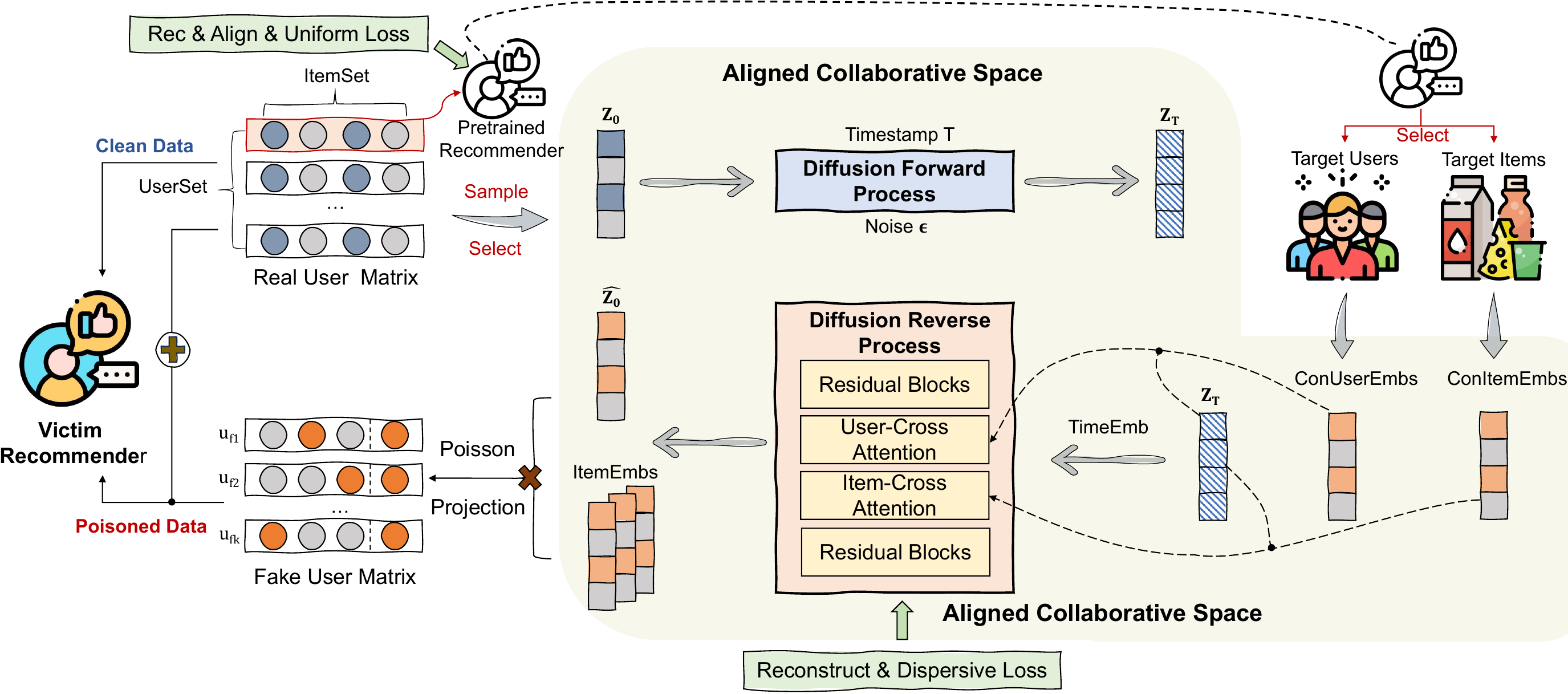}
    \caption{The framework overview of DLDA.}
    \label{fig:overview}
    
\end{figure*}

\begin{table}[t]
\small
\centering
\caption{Key Notations used throughout the paper.}
\label{tab:notation}
\resizebox{\linewidth}{!}{
\begin{tabular}{@{}ll@{}}
\toprule
\textbf{Symbol} & \textbf{Description} \\
\midrule

\multicolumn{2}{l}{\textit{General Sets and Data}} \\
$\mathcal{U}$ & Set of genuine users, $|\mathcal{U}| = M$ \\
$\mathcal{I}$ & Set of items, $|\mathcal{I}| = N$ \\
$\mathcal{U}_f$ & Set of fake users, $|\mathcal{U}_f| = M_a$ \\
$R \in \{0,1\}^{M \times N}$ & Original user-item interaction matrix \\
$\widetilde{R} \in \{0,1\}^{M_a \times N}$ & Generated fake interaction matrix \\
$R'$ & Final poisoned matrix after injection \\
$\mathcal{I}_\text{target} \subseteq \mathcal{I}$ & Target item set for promotion \\
$\text{Top@}K(u)$ & Top-$K$ recommendation list for user $u$ \\
$\mathbb{I}[\cdot]$ & Indicator function for target hit \\

\addlinespace
\multicolumn{2}{l}{\textit{Embeddings and Model}} \\
$\mathbf{e}_u$, $\mathbf{e}_i$ & User/item embedding vectors \\
$\mathbf{E}$ & Item embedding matrix \\
$\mathbf{z}_u^c$, $\mathbf{z}_v^c$ & Conditional user/item embeddings \\
$\mathcal{Z}_u$, $\mathcal{Z}_v$ & Sets of conditional embeddings \\
$L_{\text{rec}}$ & Recommendation loss for pretraining \\
$L_{\text{align}}$, $L_{\text{uniform}}$ & Alignment and uniformity losses \\

\addlinespace
\multicolumn{2}{l}{\textit{Latent Diffusion Process}} \\
$\mathbf{z}_0$, $\mathbf{z}_t$ & Latent variables (initial and noisy) \\
$\mathbf{z}'_{t}$ & Updated latent after dual cross-attention \\
$\boldsymbol{\epsilon}$ & Gaussian noise for forward process \\
$\boldsymbol{\eta}$ & Gaussian noise for reverse process \\
$t$, $\mathbf{e}_t$ & Diffusion timestep and its embedding \\
$\boldsymbol{\epsilon}_\theta$ & Denoiser network output \\

\addlinespace
\multicolumn{2}{l}{\textit{Projection to Ratings}} \\
$s$ & Item relevance score vector \\
$\mathcal{I}_{\text{active}}$ & Active item set selected for rating \\
$\lambda_\text{Pois}$ & Poisson mean for controlling rating count \\
$\delta$ & Minimum score threshold \\

\addlinespace
\multicolumn{2}{l}{\textit{Dispersive Regularization}} \\
$\mathbf{m}_i$ & Bottleneck latent representation for user $i$ \\
$\tau$ & Temperature parameter for dispersion strength \\
$\lambda_\text{disp}$ & Coefficient for dispersive loss \\
$L_{\text{diff}}$ & Diffusion reconstruction loss \\
$L_{\text{disp}}$ & Dispersive regularization loss \\
$L_{\text{total}}$ & Total training objective loss \\

\addlinespace
\multicolumn{2}{l}{\textit{Others}} \\
$\mathbf{c}$ & Centroid of user embeddings (for analysis) \\

\bottomrule
\end{tabular}
}
\end{table}

\section{Preliminaries}
\subsection{Collaborative Filtering}
CF aims to predict a user's preference for items based on their historical interactions with items. Given a set of users $\mathcal{U} = \{u_1, u_2, \dots, u_M\}$ and a set of items $\mathcal{I} = \{i_1, i_2, \dots, i_N\}$, the interaction matrix $\mathbf{R} \in \mathbb{R}^{M \times N}$ indicates whether user $u$ has interacted with item $i$, i.e., $R_{ui} = 1$ if interacted and $R_{ui} = 0$ otherwise.

The goal of CF is to learn user embeddings $\mathbf{e}_u \in \mathbb{R}^d$ and item embeddings $\mathbf{e}_i \in \mathbb{R}^d$ such that the predicted preference score $\hat{y}_{ui} = \mathbf{e}_u^\top \mathbf{e}_i$ approximates the ground truth interaction.
For personalized ranking using Bayesian Personalized Ranking (BPR) loss, the objective function is:
\[
\mathcal{L}_\text{BPR} = \sum_{(u,i,j)} \ln \sigma(\hat{y}_{ui} - \hat{y}_{uj}) + \lambda ||\Theta||^2,
\]
where \( \sigma(\cdot) \) is the sigmoid function, \( \Theta \) represents the model parameters, and \( \lambda \) is the regularization parameter.

\subsection{Targeted Shilling Attacks}
\subsubsection{Attack Goal}
A targeted shilling attack aims to increase the exposure of specific items $\mathcal{I}_{\text{target}} \subseteq \mathcal{I}$ in the top-$K$ recommendations of benign users. 

\subsubsection{Prior Knowledge}
DLDA operates under a practical black-box model assumption. The attacker has access to partial user-item interaction data, such as public logs or historical snapshots, but has no access to the parameters, architecture, or gradients of the underlying RS. Moreover, no information about user demographics or item content features is assumed. 

\subsubsection{Attacker's Capability}
The attacker can inject a small number of fake users into the RS, typically around 1\% of the normal user base. 
Each fake user's interactions are capped at the average activity level of normal users, as commonly adopted in prior studies~\cite{lin2020attacking, song2020poisonrec, nguyen2023poisoning, wang2024unveiling}.
Unlike optimization-based attacks that rely on surrogate models, DLDA operates entirely in a pre-aligned latent space, avoiding surrogate training and enhancing scalability.
To achieve this, the attacker injects $M_a$ fake users $\mathcal{U}_\text{f}$, each with a crafted interaction vector $\mathbf{r}_u \in \mathbb{R}^{|\mathcal{I}|}$. These vectors form a fake interaction matrix $\tilde{\mathbf{R}} \in \mathbb{R}^{M_a \times |\mathcal{I}|}$, which is appended to the original data:
$\mathbf{R}' = [\mathbf{R}; \tilde{\mathbf{R}}]$.
The optimization objective is to find a set of fake interactions, such that the hit rate of target items among test users are maximized:
\[
\max_{\{\mathbf{r}_u\}_{u \in \mathcal{U}_{\text{f}}}} 
\frac{1}{|\mathcal{U}_{\text{test}}|} 
\sum_{u \in \mathcal{U}_{\text{test}}} 
\mathbb{I}\big[\, \mathcal{I}_{\text{target}} \cap \text{Top@}K(u) \neq \emptyset \big].
\]

\section{Methodology}
In this section, we present DLDA, which addresses two key challenges in shilling attacks: improving the controllability of fake user behaviors toward target item promotion and preserving high stealthiness. %
Figure~\ref{fig:overview} illustrates the overall framework, and notations are listed in table~\ref{tab:notation}.

\subsection{Collaborative Latent Space Construction}

To address the dual challenges, we first construct a collaborative latent space pretrained by a CF recommender as the foundation of our approach. This space captures high-order user-item patterns and serves as a structural prior for generating profiles that are both realistic and manipulable. 
To build this space, we randomly sample 25\% of the full interaction matrix $\mathbf{R}$ to form a subset $\mathbf{R}_{\text{sample}}$, on which we pretrain a CF backbone (e.g., LightGCN) to obtain user and item embeddings.
These embeddings serve two purposes: (i) they act as conditional anchors $\mathbf{z}_u^c, \mathbf{z}_v^c$ for guiding the diffusion process, and (ii) they define the shared collaborative latent space in which fake profiles are initialized and evolved.

To further improve the expressiveness and coverage of this space, we regularize the pretraining process with an embedding consistency loss~\cite{wang2022towards}, alongside the standard recommendation objective $\mathcal{L}_{\text{rec}}$. This promotes alignment and uniformity between user and item embeddings, ensuring that both reside in a well-structured latent space. Such consistency is essential for supporting controllable and diverse generation in later stages.
\begin{equation}
\mathcal{L}_{\text{align}} = \mathbb{E}{(u,i) \sim p_{\text{pos}}} \left| f(\tilde{u}) - f(\tilde{i}) \right|^2,
\end{equation}

\begin{equation}
\begin{aligned}
\mathcal{L}_{\text{uniform}} = &\ \log \, \mathbb{E}_{u,u'\sim p_{\text{user}}} \left[ e^{-2\left\| f(\tilde{u}) - f(\tilde{u}') \right\|^2} \right] / 2 \\
&\ + \log \, \mathbb{E}_{i,i'\sim p_{\text{item}}} \left[ e^{-2\left\| f(\tilde{i}) - f(\tilde{i}') \right\|^2} \right] / 2
\end{aligned}
\end{equation}
Therefore, the loss of the pre-trained recommender is:
\begin{equation}
    \mathcal{L}_\text{pre} = \mathcal{L}_\text{rec} + \lambda_\text{au} (\mathcal{L}_{\text{align}} + \mathcal{L}_{\text{uniform}})
    \label{eq:preloss}
\end{equation}

Finally, we select high-activity genuine users as the basis for fake user initialization $\mathbf{z}_0$, so that generated profiles naturally inherit realistic interaction density and behavioral diversity, which helps improve stealthiness and reduce detectability.

\subsection{Conditional Diffusion with Dual Cross-Attention}

One of the core objectives of shilling attacks is to effectively promote specific target items, which requires precise control over the profile generation process. Traditional generative models, such as VAEs~\cite{kingma2013auto,bowman2016generating} and GANs~\cite{goodfellow2014generative,lucic2018gans}, often suffer from issues like oversmoothing or training instability, which hinder their ability to generate fine-grained and controllable user behavior patterns. To overcome these limitations, a straightforward mediation is to adopt a diffusion-based framework, which has recently demonstrated superior generative quality and controllability~\cite{rombach2022high}.

While diffusion models offer a promising alternative, existing approaches in shilling attacks remain limited in achieving precise target-oriented control. For example, TODA~\cite{liu2024toda} conditions generation solely on the target item embedding, ignoring user-specific context and leading to less personalized behaviors. To address this, we incorporate user-side conditioning, which is essential for effective target promotion (see Figure~\ref{fig:ablation}).

Building on this, we propose a latent conditional diffusion framework operating in a collaborative filtering space. At each denoising step, a dual cross-attention mechanism guides the generation: one branch attends to the target user embedding $\mathbf{z}_u^c$ to capture personalized behaviors relevant to the target item, while the other focuses on the target item $\mathbf{z}_v^c$ for effective promotion. The two signals are fused via residual addition and injected into the denoising network. This design enables the generation of fake profiles that are both target-aware and behaviorally realistic.

\paragraph{Forward Process.}
Given a timestep $t$ and an initial latent vector $\mathbf{z}_0 \in \mathbb{R}^d$, the forward process gradually corrupts $\mathbf{z}_0$ with Gaussian noise:
\begin{equation}
\mathbf{z}_t = \sqrt{\bar{\alpha}_t}\, \mathbf{z}_0 + \sqrt{1 - \bar{\alpha}_t}\, \boldsymbol{\epsilon},
\quad \boldsymbol{\epsilon} \sim \mathcal{N}(\mathbf{0}, \mathbf{I}).
\label{eq:forward}
\end{equation}

\paragraph{Reverse Process.}
A parameterized denoiser $\epsilon_\theta$ predicts and removes noise:
\begin{multline}
\mathbf{z}_{t-1} = 
\frac{1}{\sqrt{\alpha_t}}
\Big(
\mathbf{z}_t - 
\frac{\beta_t}{\sqrt{1 - \bar{\alpha}_t}}\,
\epsilon_\theta(\mathbf{z}_t, t, \mathbf{z}_u^c, \mathbf{z}_v^c)
\Big) \\
+ \mathbb{I}_{t>0}\, \sqrt{\beta_t}\, \boldsymbol{\eta},
\quad \boldsymbol{\eta} \sim \mathcal{N}(\mathbf{0}, \mathbf{I}).
\label{eq:denoiser}
\end{multline}

\paragraph{Denoiser Architecture.}
The denoiser adopts a U-Net backbone with the following components:
\begin{itemize}[leftmargin=*]
  \item \textbf{Residual Blocks:} Each block preserves information via skip connections:
  \begin{equation}
  \mathbf{h}^{(l+1)} = \mathbf{h}^{(l)} + \mathrm{MLP}\big(\mathbf{h}^{(l)}\big).
  \end{equation}
  \item \textbf{Dual Cross-Attention:} For each $t$, the latent query $\mathbf{q} = \mathbf{z}_t$ attends to both conditional embeddings:
\begin{align}
\mathbf{a}_u &= \text{CrossAttn}(\mathbf{q},\, \mathbf{z}_u^c), 
\quad
\mathbf{a}_v = \text{CrossAttn}(\mathbf{q},\, \mathbf{z}_v^c),\\
\mathbf{z}_t' &= \mathbf{z}_t + \mathbf{a}_u + \mathbf{a}_v.
\label{eq:crossattn}
\end{align}

  \item \textbf{Time Embedding:} The timestep $t$ is encoded using sinusoidal position embeddings and projected to match the latent dimension.
\end{itemize}

\paragraph{Training Objective.}
At inference, the generator starts from Gaussian noise and iteratively applies the learned reverse process to sample a clean latent vector $\mathbf{z}_0^{\text{fake}}$, which is decoded into a sparse rating profile.
The training objective minimizes the noise prediction error under conditional guidance:
\begin{equation}
\mathcal{L}_{\text{diff}} = 
\mathbb{E}_{\mathbf{z}_0,\, t,\, \boldsymbol{\epsilon}} 
\Big[
\big\|
\epsilon_\theta\big(
\sqrt{\bar{\alpha}_t}\, \mathbf{z}_0 + \sqrt{1-\bar{\alpha}_t}\, \boldsymbol{\epsilon},\,
t,\, \mathbf{z}_u^c,\, \mathbf{z}_v^c
\big)
- \boldsymbol{\epsilon}
\big\|_2^2
\Big],
\label{eq:diff}
\end{equation}
where $\mathbf{z}_0$ is initialized from the pretrained recommender, $t$ is uniformly sampled from $\{1,\dots,T\}$, and $\boldsymbol{\epsilon}$ is standard Gaussian noise.
This conditional objective ensures that the denoiser learns to recover realistic latent representations aligned with both the target user and target item characteristics. The overall training process—including conditional noising, denoising, and loss computation is implemented in Algorithm~\ref{alg:dlda}, Lines 5-7.

\subsection{Dispersive Regularization }

Another key objective of shilling attacks is to ensure stealthiness, which requires generating diverse yet realistic user behaviors that naturally blend into the genuine population. However, existing generative attacks that prioritize target exposure often produce concentrated fake profiles, forming detectable behavioral clusters. To mitigate this, we introduce a \emph{Dispersive Loss} on intermediate latent features to promote diversity and prevent mode collapse during generation. 
Inspired by recent work~\cite{wang2025diffuse} showing that dispersion among hidden states improves sample variety without contrastive pairs, we apply this loss to the bottleneck features $\mathbf{m}_i \in \mathbb{R}^d$ from our UNet-style denoiser~\cite{rombach2022high}, defined as:
\begin{equation}
\mathcal{L}_{\text{disp}} = \log \mathbb{E}_{i \neq j} \Big[ 
\exp \Big( - \frac{\|\mathbf{m}_i - \mathbf{m}_j\|_2^2}{\tau} \Big) 
\Big],
\label{eq:disp}
\end{equation}
where $\tau$ is a temperature hyperparameter controlling the repulsion strength.

This regularization acts as a soft repulsive force that disperses latent representations, promoting intra-group diversity while maintaining alignment with the real user manifold. This mitigates local mode collapse and enhances stealthiness by preventing tight, detectable fake user clusters (see Appendix~\ref{appendix: B} for an information-theoretic intuition).

We combine this regularization with the standard diffusion reconstruction loss to form the final objective:
\begin{equation}
\mathcal{L}_{\text{total}} = \mathcal{L}_{\text{diff}} + \lambda_{\text{disp}} \cdot \mathcal{L}_{\text{disp}},
\label{eq:total}
\end{equation}
where $\lambda_{\text{disp}}$ controls the trade-off between reconstruction fidelity and behavioral diversity.
Eq.~\ref{eq:total} defines the full training objective, which is optimized during the denoising updates in Algorithm~\ref{alg:dlda}, Line 7.

\subsection{Poisson-Based Rating Projection}

The diffusion model outputs a continuous latent vector $\mathbf{z}_0^{\text{fake}}$, which must be projected into a sparse binary rating vector to mimic real-world user behavior. Prior attacks~\cite{lin2020attacking, rong2022poisoning, wang2024unveiling} typically apply fixed-size top-$k$ selection, overlooking the natural variability in user activity levels. In contrast, we adopt a Poisson-based strategy to generate variable-length interactions.

We first compute item relevance scores by projecting the latent vector into the item embedding space:
\begin{equation}
\mathbf{s} = \mathbf{E} \cdot \mathbf{z}_0^{\text{fake}} \in \mathbb{R}^N,
\label{eq:score_proj}
\end{equation}
where $\mathbf{E} \in \mathbb{R}^{N \times d}$ is the pretrained item embedding matrix.

Next, we sample the number of interactions from a Poisson distribution:
\begin{equation}
n \sim \text{Poisson}(\lambda_\text{Pois}), \quad n = \min(n, N, n_{\text{max}}),
\end{equation}
and select the top-$n$ items with scores above a threshold:
\begin{equation}
\mathcal{I}_{\text{active}} = \left\{ i \in \text{Top-}n(\mathbf{s}) \mid \mathbf{s}_i \geq \delta \right\}.
\end{equation}

We then construct the binary rating vector as:
\begin{equation}
\tilde{\mathbf{r}}[i] =
\begin{cases}
1 & \text{if } i \in \mathcal{I}_{\text{active}} \cup \mathcal{I}_\text{target} \\
0 & \text{otherwise}
\end{cases}.
\label{eq:binary_proj}
\end{equation}

This projection ensures both realism and controllability by aligning with user activity patterns while enforcing target exposure. The full procedure is implemented in Algorithm~\ref{alg:dlda}, Lines 10-11, completing the DLDA generation pipeline.

\begin{algorithm}[t]
\caption{DLDA}
\label{alg:dlda}

\begin{minipage}{\linewidth}
\begin{algorithmic}[1]
\REQUIRE Pretrained recommender, interaction matrix $\mathbf{R}$, target item set $\mathcal{I}_\text{target}$, number of fake users $M_a$
\ENSURE Poisoned matrix $\tilde{\mathbf{R}}$
\STATE Pretrain recommender on $\mathbf{R}_{\text{sample}}$ and extract embeddings $\mathbf{Z}_u, \mathbf{Z}_v$ with $\mathcal{L}_\text{pre}$ in Eq.~\ref{eq:preloss}
\STATE Select condition users $\mathcal{U}_{\text{cond}}$ interacting with $\mathcal{I}_\text{target}$
\STATE Construct $\mathbf{Z}_u^c$, $\mathbf{Z}_v^c$ from user/item embeddings
\FOR{epoch = 1 to $E$}
    \STATE Sample genuine users and generate latent $\mathbf{z}_t$ by noising (Eq.~\ref{eq:forward})
    \STATE Predict noise with generator $\mathcal{G}_\theta$ and compute loss using conditional denoising (Eq.~\ref{eq:denoiser}, \ref{eq:crossattn}, \ref{eq:diff})
    \STATE Update $\theta$ with diffusion + dispersive loss (Eq.~\ref{eq:diff}, \ref{eq:disp}, \ref{eq:total})
\ENDFOR
\FOR{each fake user $i=1$ to $M_a$}
    \STATE Sample $\mathbf{z}_T \sim \mathcal{N}(0, I)$ and denoise via $\mathcal{G}_\theta$
    \STATE Generate scores and project via Poisson strategy (Eq.~\ref{eq:score_proj}-\ref{eq:binary_proj})
    \STATE Inject positive ratings on target items $\mathcal{I}_\text{target}$
\ENDFOR
\RETURN $\mathbf{R}^{'} = \mathbf{R} \cup \{\tilde{\mathbf{r}}_i\}_{i=1}^{M_a}$
\end{algorithmic}
\end{minipage}

\end{algorithm}

\section{Experiments}
In this section, we conduct experiments to answer the following research questions (RQs):
\begin{itemize}[leftmargin=*]
    \item \textbf{RQ1.} How effective is DLDA in promoting target items under different recommendation models?
    \item \textbf{RQ2.} How does each component affect DLDA’s performance?
    \item \textbf{RQ3.} How does the dispersive weight impact the trade-off between success and stealth?
    \item \textbf{RQ4.} How detectable are DLDA-generated users?
    \item \textbf{RQ5.} How efficient is DLDA in time and space?
\end{itemize}
\subsection{Experiment Setup}
\subsubsection{Datasets}

\begin{table}[t]
    \caption{Statistics of datasets.}
    \centering
    \resizebox{0.9\linewidth}{!}{
    \begin{tabular}{c|cccccc} \hline
    \textbf{Datasets}   & \textbf{Users}  & \textbf{Items} & \textbf{Interactions}   & \textbf{Avg.Int.} &\textbf{{Sparsity(\%)}}  \\ \hline
    ML-100K & 942  & 1,447  & 55,375   & 38.27 & 96.49  \\
    ML-1M & 6,038  & 3,533  & 575,281 & 162.83 & 97.59 \\ 
    Douban & 2,848  & 39,586  & 894,887 & 22.61 & 99.23 \\ 
    \hline
    \end{tabular}
    }
    \label{tab:dataset_statistics}
\end{table}

To comprehensively evaluate the effectiveness and robustness of our attack model, we conduct experiments on three real-world recommendation datasets with varying levels of sparsity: MovieLens-100K (ML-100K)\cite{fang2018poisoning}, MovieLens-1M (ML-1M), and Douban\cite{zhao2016user}. These datasets range from relatively dense (ML-100K) to highly sparse (Douban), facilitating a thorough investigation into the transferability and stealthiness of the attack across diverse recommendation environments. Following the preprocessing protocol of CLeaR \cite{wang2024unveiling}, we prepare the datasets accordingly, and the detailed statistics are summarized in Table \ref{tab:dataset_statistics}.

\begin{table*}[t]
\centering
\caption{Comparison of unpopular target promotion performance across attack methods on ML-100K, ML-1M, and Douban datasets. Bold indicates the best, and underline indicates the second best.}
\label{tab:unpopular}
\resizebox{\textwidth}{!}{
\begin{tabular}{llcccccccccccccccc} 
\toprule
\textbf{Dataset} & \textbf{Model} & \multicolumn{2}{c}{None} & \multicolumn{2}{c}{Random} & \multicolumn{2}{c}{Bandwagon} & \multicolumn{2}{c}{AUSH} & \multicolumn{2}{c}{GSPAttack} & \multicolumn{2}{c}{CLeaR} & \multicolumn{2}{c}{DLDA} & \multicolumn{2}{c}{Improv. (\%)} \\
\cmidrule(lr){3-4} \cmidrule(lr){5-6} \cmidrule(lr){7-8} \cmidrule(lr){9-10}
\cmidrule(lr){11-12} \cmidrule(lr){13-14} \cmidrule(lr){15-16} \cmidrule(lr){17-18}
& & H@50 & N@50 & H@50 & N@50 & H@50 & N@50 & H@50 & N@50 & H@50 & N@50 & H@50 & N@50 & H@50 & N@50 & H@50 & N@50 \\
\midrule

\multirow{5}{*}{ML-100K} 
& NGCF      & 0.0057 & 0.0019 & 0.0180 & 0.0070 & 0.0142 & 0.0056 & 0.0177 & 0.0070 & 0.0172 & 0.0081 & \underline{0.0199} & \underline{0.0080} & \textbf{0.0206} & \textbf{0.0090} & +3.51 & +12.50 \\
& LightGCN  & 0.0101 & 0.0037 & 0.0170 & 0.0068 & \underline{0.0178} & 0.0075 & 0.0158 & 0.0063 & 0.0168 & 0.0069 & 0.0175 & \underline{0.0075} & \textbf{0.0220} & \textbf{0.0094} & +23.59 & +25.33 \\
& SimGCL    & 0.0290 & 0.0144 & 0.0368 & 0.0175 & \underline{0.0370} & \underline{0.0180} & 0.0353 & 0.0170 & 0.0362 & 0.0172 & 0.0366 & 0.0174 & \textbf{0.0389} & \textbf{0.0181} & +5.13 & +0.55 \\
& EGCF      & 0.0269 & 0.0104 & 0.0256 & 0.0098 & 0.0248 & 0.0097 & 0.0280 & 0.0109 & \underline{0.0283} & \underline{0.0110} & 0.0278 & 0.0107 & \textbf{0.0295} & \textbf{0.0115} & +4.24 & +4.54 \\
& LightCCF  & 0.0343 & 0.0167 & 0.0346 & 0.0173 & 0.0349 & 0.0177 & 0.0353 & 0.0176 & 0.0358 & 0.0182 & \underline{0.0363} & \underline{0.0184} & \textbf{0.0367} & \textbf{0.0188} & +1.10 & +2.17 \\
\midrule

\multirow{5}{*}{ML-1M} 
& NGCF      & 0.0004 & 0.0002 & 0.0021 & 0.0042 & 0.0020 & 0.0030 & 0.0032 & \underline{0.0082} & \underline{0.0039} & 0.0077 & 0.0020 & 0.0047 & \textbf{0.0083} & \textbf{0.0099} & +112.82 & +20.73 \\
& LightGCN  & 0.0002 & 0.0001 & 0.0022 & 0.0070 & 0.0023 & 0.0089 & \underline{0.0026} & \underline{0.0096} & 0.0022 & 0.0074 & 0.0023 & 0.0055 & \textbf{0.0030} & \textbf{0.0112} & +15.38 & +16.66 \\
& SimGCL    & 0.0001 & 0.0000 & 0.0034 & 0.0097 & 0.0030 & 0.0084 & \underline{0.0036} & \textbf{0.0099} & 0.0020 & 0.0072 & 0.0032 & 0.0056 & \textbf{0.0039} & \textbf{0.0099} & +8.33 & +0.00 \\
& EGCF      & 0.0005 & 0.0002 & 0.0017 & 0.0022 & 0.0031 & 0.0071 & \underline{0.0037} & 0.0085 & 0.0019 & 0.0066 & 0.0021 & \textbf{0.0093} & \textbf{0.0043} & \underline{0.0087} & +7.50 & -6.45 \\
& LightCCF  & 0.0001 & 0.0001 & 0.0023 & 0.0051 & 0.0034 & 0.0078 & \underline{0.0057} & \underline{0.0090} & 0.0019 & 0.0016 & 0.0023 & 0.0025 & \textbf{0.0059} & \textbf{0.0096} & +3.51 & +6.67 \\
\midrule

\multirow{5}{*}{Douban} 
& NGCF      & 0.0002 & 0.0001 & 0.0033 & 0.0044 & 0.0023 & 0.0016 & 0.0026 & 0.0047 & \underline{0.0040} & \underline{0.0056} & 0.0028 & 0.0028 & \textbf{0.0080} & \textbf{0.0068} & +100.00 & +21.42 \\
& LightGCN  & 0.0001 & 0.0000 & 0.0009 & 0.0010 & 0.0022 & 0.0015 & 0.0025 & 0.0017 & 0.0025 & 0.0016 & \underline{0.0026} & \underline{0.0020} & \textbf{0.0029} & \textbf{0.0045} & +11.53 & +125.00 \\
& SimGCL    & 0.0001 & 0.0000 & 0.0108 & 0.0320 & 0.0110 & 0.0304 & \underline{0.0178} & \underline{0.0552} & 0.0060 & 0.0171 & 0.0048 & 0.0114 & \textbf{0.0212} & \textbf{0.0621} & +19.10 & +12.50 \\
& EGCF      & 0.0000 & 0.0000 & 0.0003 & 0.0008 & 0.0052 & 0.0162 & \underline{0.0070} & \underline{0.0255} & 0.0059 & 0.0131 & 0.0025 & 0.0098 & \textbf{0.0182} & \textbf{0.0539} & +160.00 & +111.37 \\
& LightCCF  & 0.0002 & 0.0001 & 0.0004 & 0.0002 & 0.0046 & 0.0121 & 0.0058 & 0.0143 & \underline{0.0073} & \underline{0.0179} & 0.0013 & 0.0018 & \textbf{0.0091} & \textbf{0.0220} & +24.65 & +22.90 \\
\bottomrule
\end{tabular}}
\label{tab:unpopular}
\end{table*}

\subsubsection{Implementation Details and Hyperparameter Settings}
All experiments are conducted on NVIDIA A40 GPUs using PyTorch 2.1.0 and Python 3.9 under a managed conda environment. Datasets are split into training/validation/test sets with an 8:1:1 ratio. Victim models are trained on the full dataset, while attack models are trained on a randomly sampled 25\% subset to simulate a black-box setting.

Poisoned data is injected only during training and does not affect validation or testing directly. Each attack is repeated five times, and average results are reported. All models use Adam optimizer with a unified learning rate of 0.005, batch size of 64, and hidden dimension of 64. The injection ratio is fixed at 1\% of users, and 5 target items are selected. Key hyperparameters are set as $\lambda_{\text{AU}}{=}0.1$, $\lambda_{\text{dist}}{=}0.5$, and $\tau{=}0.5$.

\subsubsection{Evaluation Metrics}

Target Metrics: We report Hit Ratio@K (H@K) and Normalized Discounted Cumulative Gain@K (N@K) on the target item set to quantify the direct effectiveness of target item promotion.

Global Metrics: Instead of reporting absolute values, we measure the relative change in H@K and N@K by comparing the recommender’s performance before and after poisoning, using the clean-trained model as the baseline.

\subsubsection{Victim Recommender Systems}

We evaluate our attack on a range of representative CF-based recommenders, including NGCF\cite{he2017neural}, LightGCN\cite{he2020lightgcn}, SimGCL\cite{yu2022graph}, EGCF\cite{zhang2024simplify}, and LightCCF\cite{zhang2025unveiling}. These models cover graph-based, contrastive, and embedding-free paradigms. Detailed descriptions are provided in Section~\ref{sec:related}.

\subsubsection{Shilling Attack Baselines}
We compare DLDA against representative shilling attack methods spanning heuristic, optimization-based, and generative paradigms. Specifically, we include:  
\begin{itemize}[leftmargin=*]
\item NoneAttack: No attack; serves as a clean performance baseline.
\item RandomAttack\cite{lam2004shilling}: Generates rating vectors randomly, used for robustness testing.
\item BandWagonAttack\cite{gunes2014shilling}: Heuristic method that injects the target alongside popular items.
\item AUSH\cite{lin2020attacking}: GAN-based method that generates fake profiles to promote the target item.
\item GSPAttack\cite{nguyen2023poisoning}: Optimizes fake behaviors via GNN-based gradient updates to perturb the user–item graph.
\item CLeaR\cite{wang2024unveiling}: A white-box two-stage attack combining generation and recommender-aware optimization.
\end{itemize}

\begin{table}[htbp]
    \caption{Comparison of DLDA, CLeaR, and GSPAttack on global/target H@50 and N@50 for popular item attacks on ML-100K and ML-1M.}
    \label{tab:popular}
    \centering
    \resizebox{1.0\linewidth}{!}{
    \begin{tabular}{clcccc}
        \toprule
        \textbf{Dataset} & \textbf{Model} & \textbf{Global H@50 $\Delta$} & \textbf{Target H@50} & \textbf{Global N@50 $\Delta$} & \textbf{Target N@50} \\
        \midrule
        \multirow{3}{*}{ML-100K} 
        & GSPAttack & -0.0024 & 0.1637 & 0.0141 & 0.1112 \\
        & CLeaR     &  0.0051 & 0.1622 & 0.0164 & 0.1138 \\
        & DLDA      & \textbf{0.0071} & \textbf{0.1667} & \textbf{0.0149} & \textbf{0.1276} \\
        \midrule
        \multirow{3}{*}{ML-1M} 
        & GSPAttack & -0.0047 & 0.0999 & -0.0021 & 0.0598 \\
        & CLeaR     & -0.0044 & 0.0941 & -0.0021 & 0.0630 \\
        & DLDA      & \textbf{-0.0038} & \textbf{0.1006} & \textbf{0.0003} & \textbf{0.0775} \\
        \bottomrule
    \end{tabular}
    }
    \label{tab:popular}
\end{table}

\subsection{Attack Performance Comparison (RQ1)}
We compare DLDA with baselines on three real-world datasets. Results for unpopular and popular item attacks are shown in Table~\ref{tab:unpopular} and Table~\ref{tab:popular}, respectively.

Attacking unpopular items is inherently more challenging due to limited interaction signals and weak collaborative effects. Despite this, DLDA consistently achieves the highest target exposure across all backbone recommenders. On ML-1M with NGCF, DLDA improves target H@50 by 112.82 \% over the strongest baseline, highlighting its capability to synthesize highly effective adversarial profiles even under sparse feedback. On the extremely sparse Douban dataset, DLDA still delivers significant gains, including a 160.00 \% increase in H@50 for EGCF, demonstrating strong adaptability across data regimes.

We further evaluate DLDA on promoting popular items, which are reinforced by dense interactions and typically pose higher detection risks. As shown in Table~\ref{tab:popular}, DLDA obtains the best target H@50 and N@50 on both ML-100K and ML-1M, while preserving or slightly improving global recommendation quality. On ML-100K, for instance, DLDA achieves a target H@50 of 0.1667, surpassing GSPAttack and CLeaR, and yields the highest global H@50 improvement of 0.0071. These results confirm DLDA’s ability to manipulate the ranking of widely recommended items without compromising overall system utility.

DLDA is explicitly designed to balance attack effectiveness and stealth. Leveraging latent diffusion and dispersive regularization, it generates fake user profiles that are collaboratively consistent and behaviorally diverse. This mitigates the anomaly patterns often induced by gradient-based methods and enhances real-world applicability. The trade-off is further validated by our ablation results in Section~\ref{sec:Ablation} and detection analyses in Section~\ref{sec:detection}.

\begin{figure*}[t]
    \centering
    \begin{subfigure}[b]{0.32\linewidth}
        \centering
        \includegraphics[width=\linewidth]{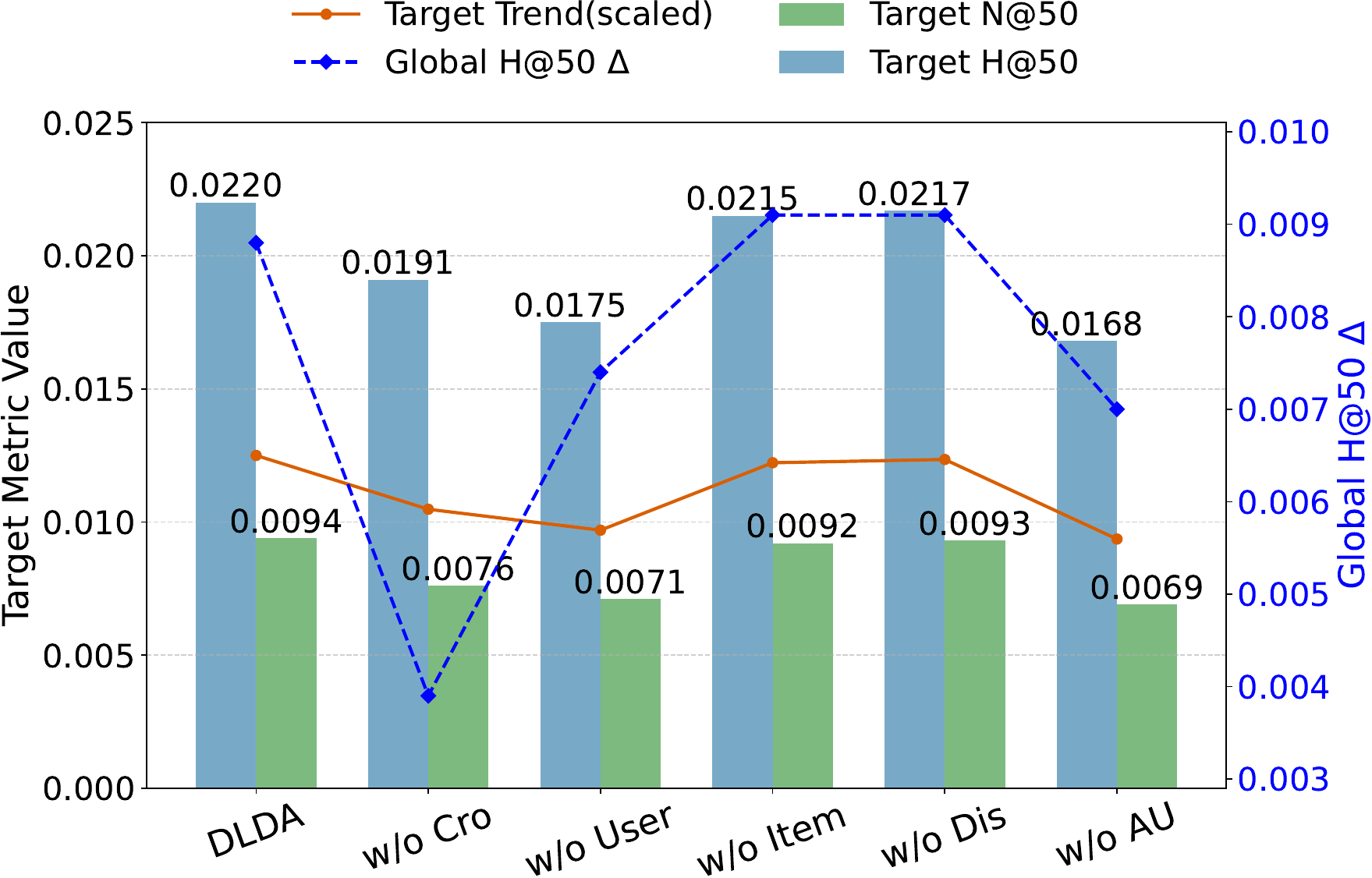}
        \caption{ML-100K}
        \label{fig:ablation_ml100k}
    \end{subfigure}
    \hfill
    \begin{subfigure}[b]{0.32\linewidth}
        \centering
        \includegraphics[width=\linewidth]{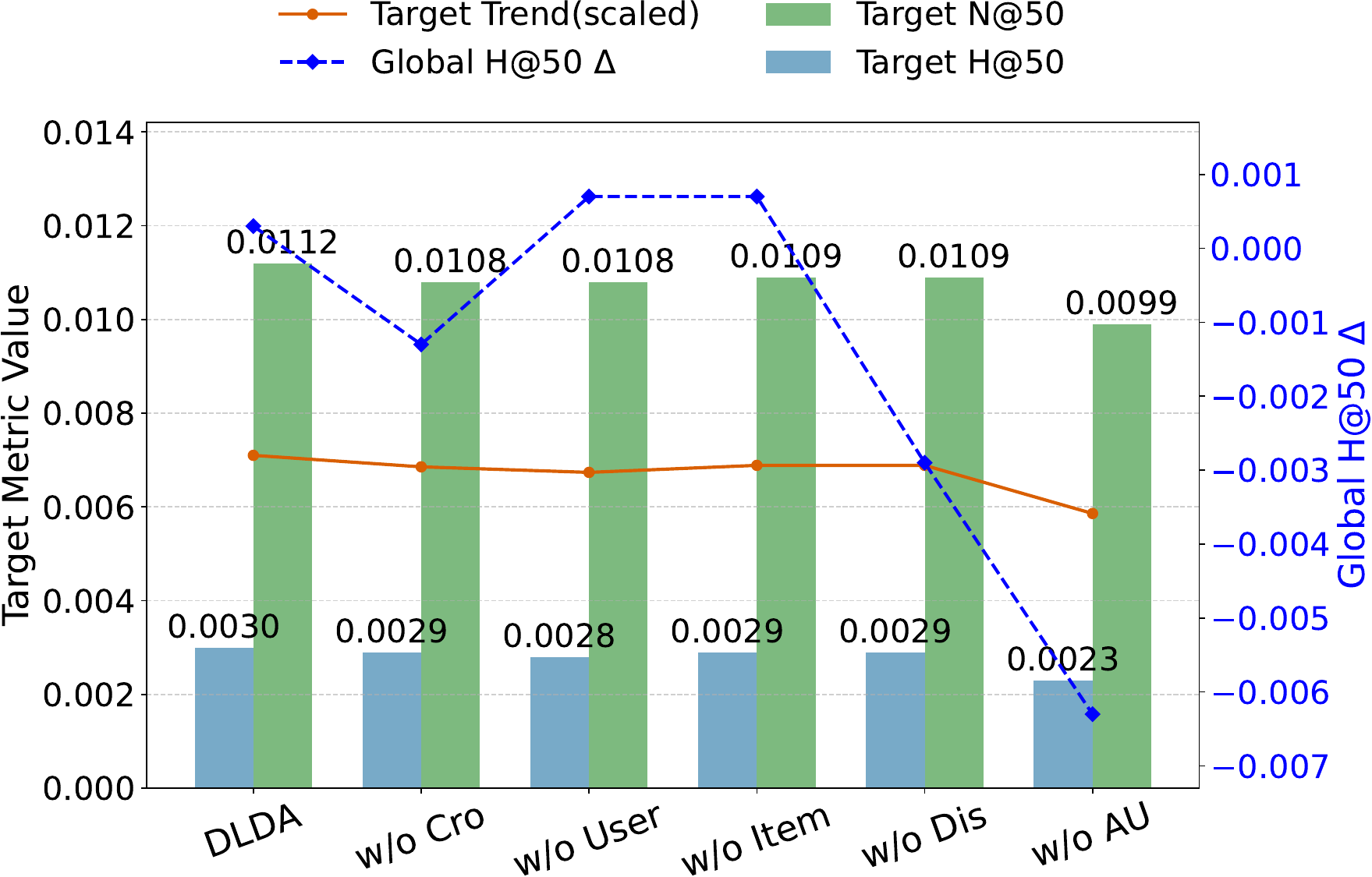}
        \caption{ML-1M}
        \label{fig:ablation_douban}
    \end{subfigure}
    \hfill
    \begin{subfigure}[b]{0.32\linewidth}
        \centering
        \includegraphics[width=\linewidth]{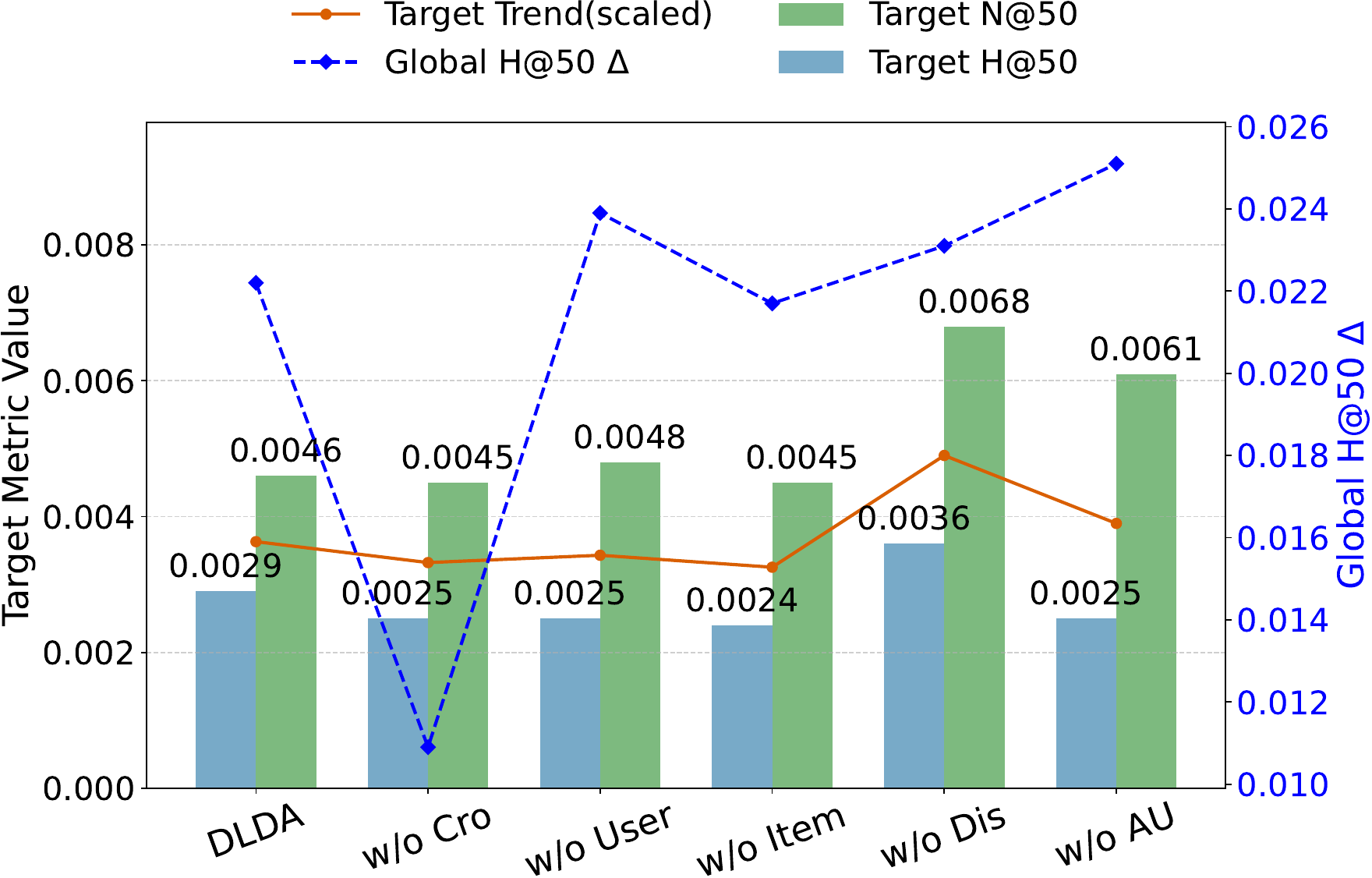}
        \caption{Douban}
        \label{fig:ablation_douban}
    \end{subfigure}
    \caption{Ablation results on three datasets.}
    \label{fig:ablation}
\end{figure*}

\begin{figure*}[t]
    \centering
    \begin{subfigure}[b]{0.32\linewidth}
        \centering
        \includegraphics[width=\linewidth]{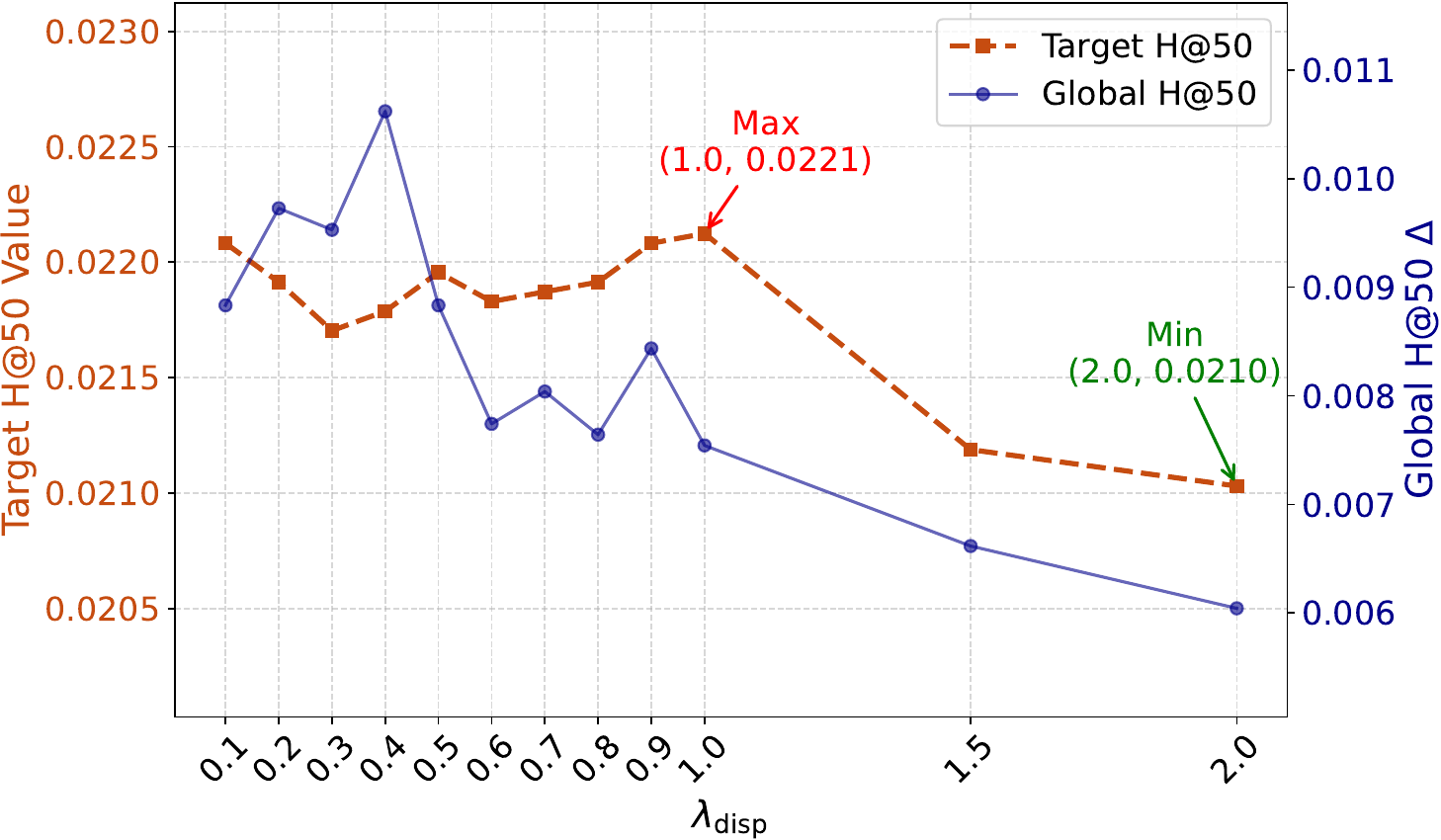}
        \caption{ML-100K.}
        \label{fig:Hyper_ml100k}
    \end{subfigure}
    \hfill
    \begin{subfigure}[b]{0.32\linewidth}
        \centering
        \includegraphics[width=\linewidth]{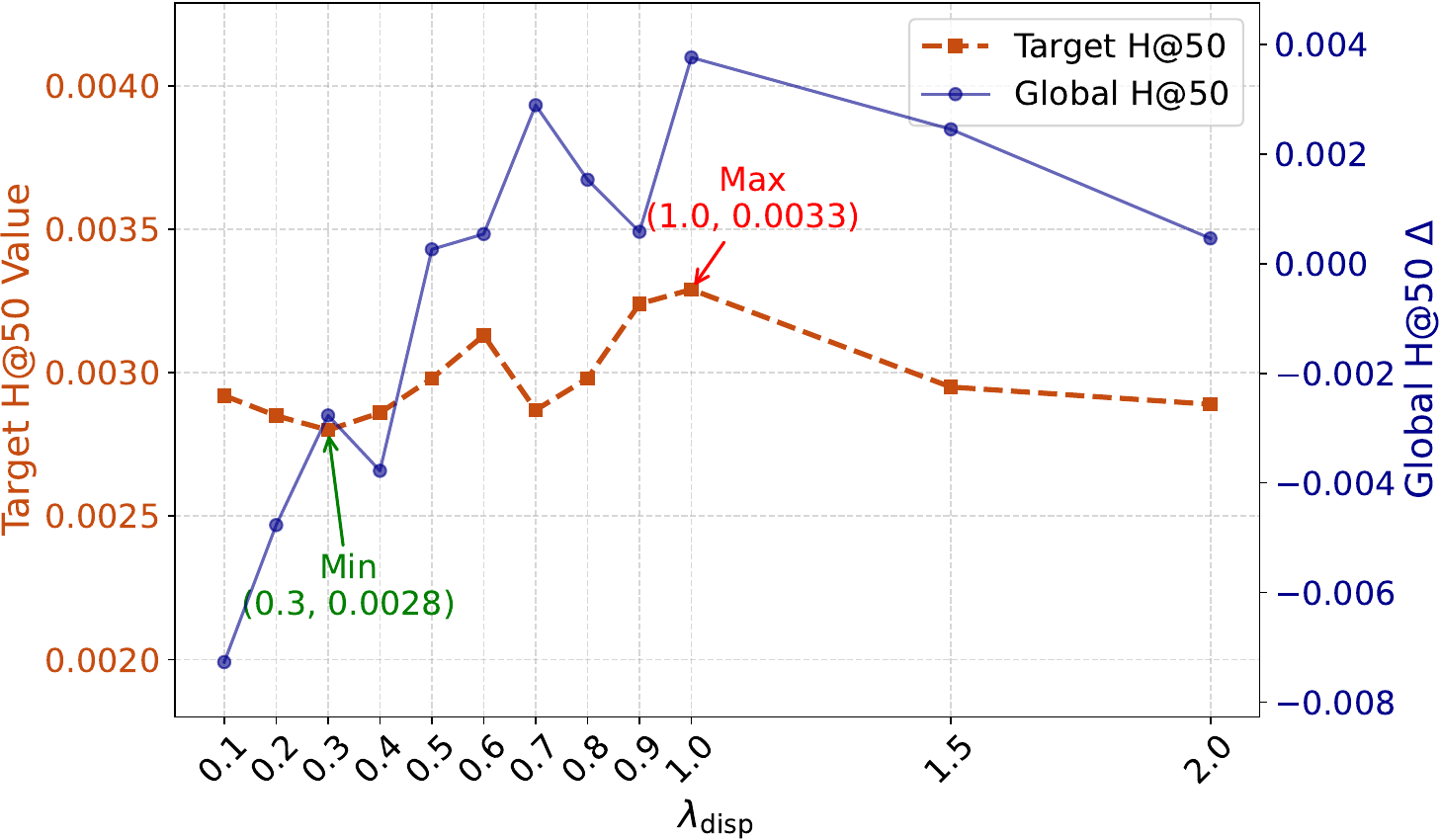}
        \caption{ML-1M.}
        \label{fig:Hyper_ml1m}
    \end{subfigure}
    \hfill
    \begin{subfigure}[b]{0.32\linewidth}
        \centering
        \includegraphics[width=\linewidth]{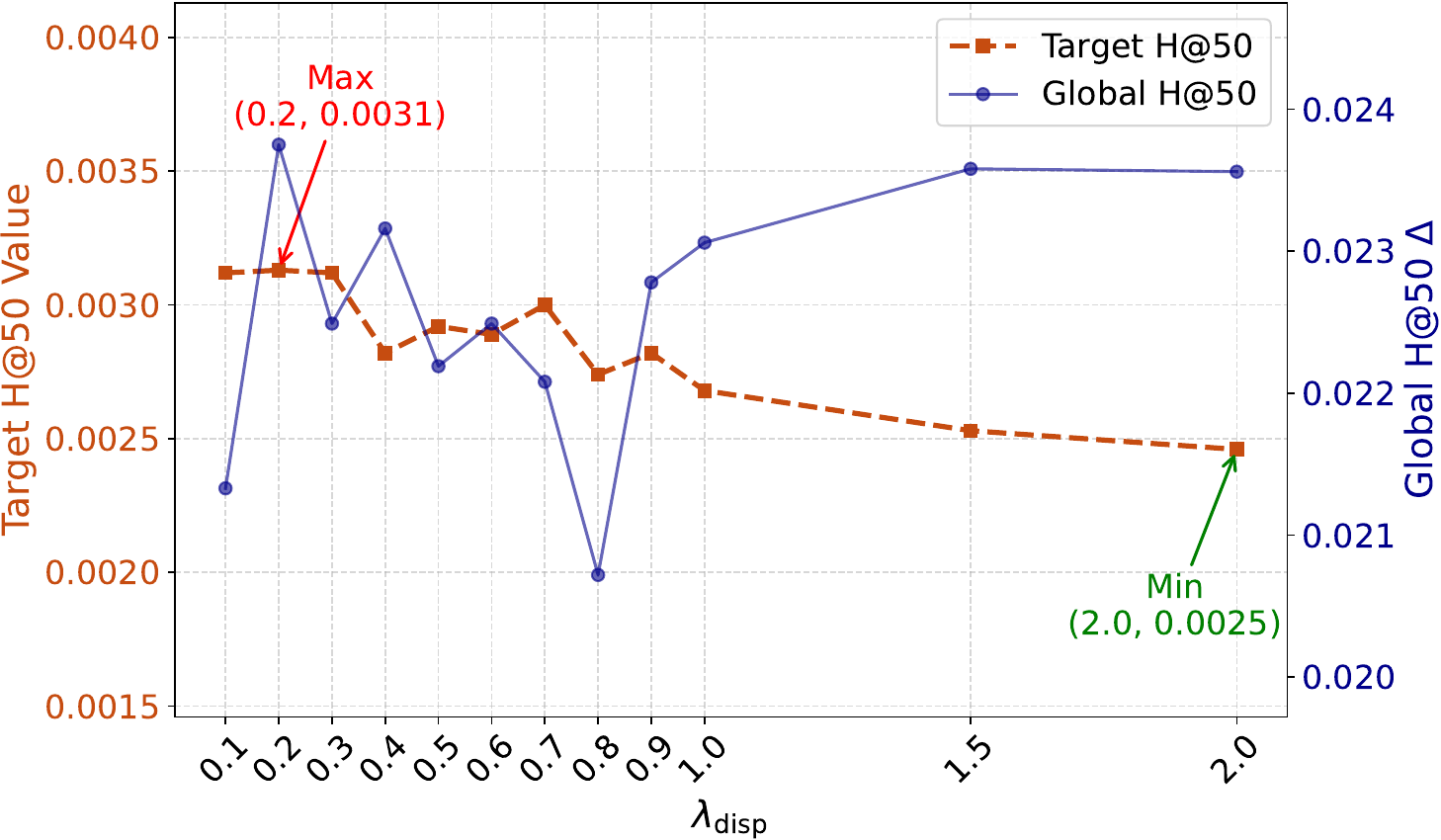}
        \caption{Douban.}
        \label{fig:Hyper_douban}
    \end{subfigure}
    \caption{Sensitivity of hyperparameter $\lambda_{\text{disp}}$ on three datasets.}
    \label{fig:hyper}
\end{figure*}

\subsection{Ablation Study (RQ2)}
\label{sec:Ablation}

To evaluate the contribution of each component in DLDA, we perform an ablation study by individually removing the following
modules: (i) Dual Cross-Attention (entirely or partially), (ii) Dispersive Loss, and (iii) Alignment-Uniformity (AU) Loss. Results on ML-100K, ML-1M, and Douban are summarized in Figure \ref{fig:ablation}.

\textbf{Dual Cross-Attention (w/o Cro, w/o User, w/o Item).} 
Removing the full dual cross-attention mechanism (w/o Cro) leads to a clear degradation in both target H@50 and N@50 across all datasets. The impact is especially strong on ML-100K, where richer user-item interactions enable more effective conditioning. Further ablations show that removing either the user (w/o User) or item (w/o Item) guidance independently also reduces performance, indicating that both sides of the guidance are necessary for controllable and target-effective generation.

\textbf{Dispersive Loss (w/o Dis).}
On ML-100K and ML-1M, removing the dispersive loss leads to noticeable drops in target H@50 (from 0.02196 to 0.02170 on ML-100K, and from 0.00298 to 0.00293 on ML-1M), despite slight improvements in global H@50. This indicates that the regularization effectively disperses fake users in the latent space, which not only enhances behavioral diversity but also improves the precision of target-oriented generation.
In contrast, on Douban, where interactions are significantly sparser, removing the dispersive loss yields the highest target H@50 (0.00360) and N@50 (0.00740), as fake users collapse toward more concentrated patterns that intensify the attack. However, this comes at the expense of stealth, reflected by a sharp decline in global H@50.

\textbf{AU Loss (w/o AU).}
While AU loss improves latent space alignment on denser datasets, its benefit diminishes under high sparsity. On Douban, removing it unexpectedly improves both target N@50 and global H@50, suggesting that strong alignment constraints may hinder attack flexibility in sparse environments.

In summary, each component of DLDA plays a distinct, complementary role, jointly ensuring that fake users are target-driven, diverse, and realistic.

\subsection{Hyperparameter Sensitivity Analysis (RQ3)}

We examine how $\lambda_{\text{disp}}$ influences DLDA’s trade-off across all datasets, as shown in Figure~\ref{fig:hyper}.

On ML-100K, increasing $\lambda_{\text{disp}}$ from 0.1 to 1.0 steadily boosts the target H@50, reaching its peak value of 0.0221 at $ \lambda_{\text{disp}} = 1.0$. However, further increase leads to a decline, dropping to 0.0210 at $\lambda_{\text{disp}} = 2.0$. The global H@50 change remains modest (within ±0.003), indicating that the added dispersion promotes intra-group diversity without overly disturbing the global ranking.

On ML-1M, a similar trend is observed. The target H@50 improves with larger $\lambda_{\text{disp}}$, peaking at 0.0033 when $\lambda_{\text{disp}} = 1.0$, and then slightly decreases beyond this point. The global H@50 variation remains nearly flat throughout, suggesting robustness to moderate regularization.

In contrast, Douban exhibits an earlier turning point: the target H@50 peaks at 0.0031 around $ \lambda_{\text{disp}} = 0.2$, followed by a continuous decline. This indicates that Douban, probably because of its higher sparsity, is more sensitive to overdispersion. Over-regularization may push fake users too far from the real-user manifold, weakening their impact.

Overall, the optimal $\lambda_{\text{disp}}$ is dataset-dependent: moderate values (0.8-1.0) suit denser datasets like ML-100K and ML-1M, while smaller ones (0.2-0.3) work better for sparse data like Douban. DLDA performs robustly across this range, requiring minimal hyperparameter tuning.
\begin{figure}[t]
    \centering
    \begin{subfigure}[b]{0.52\linewidth}
        \centering
        \includegraphics[width=\linewidth]{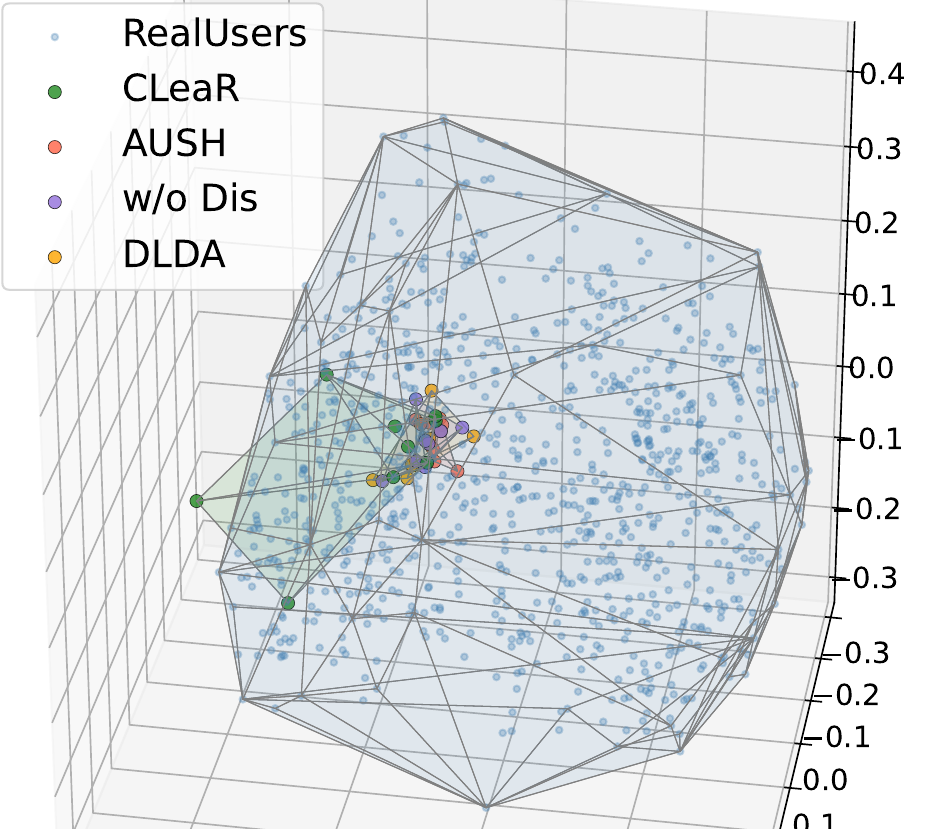}
        \caption{3D PCA shows that fake users maintain diversity while aligning with the real user manifold.}
        \label{fig:pca}
    \end{subfigure}
    \hfill
    \begin{subfigure}[b]{0.42\linewidth}
        \centering
        \includegraphics[width=\linewidth]{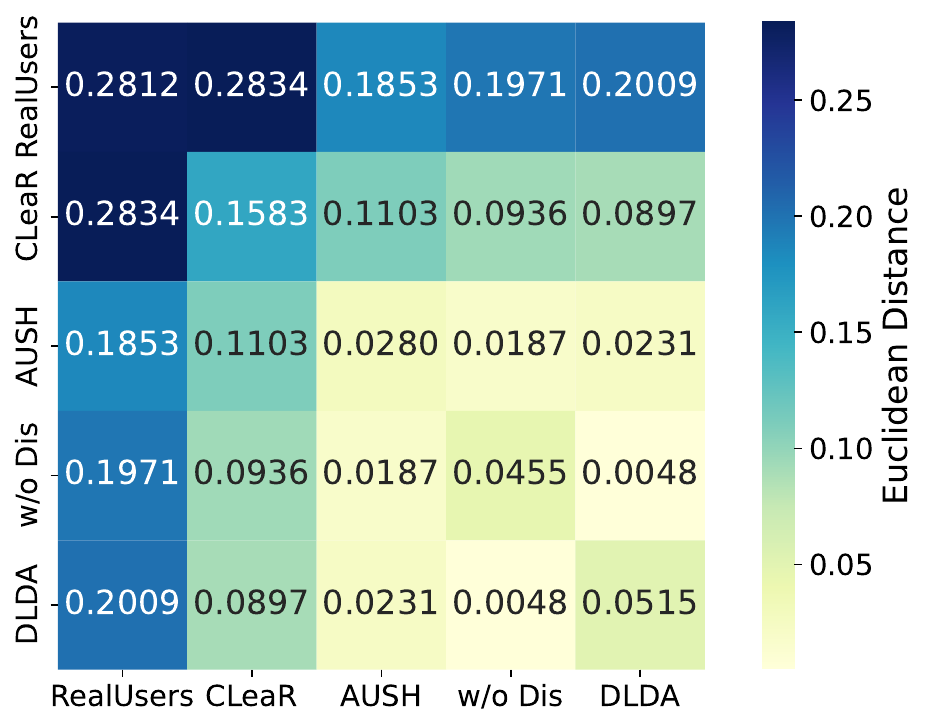}
        \caption{Heatmap of group centroid distances shows fake users blend naturally with real user clusters.}
        \label{fig:distance}
    \end{subfigure}
    \caption{Embedding visualization of attack methods on ML\textit{-}100K.}
    \label{fig:pca_distance_combined}
    
\end{figure}

\subsection{Detection \& Visualization Analytics (RQ4)}
\label{sec:detection}

\begin{table*}[htbp]
\centering
\caption{Comprehensive evaluation of stealthiness across statistical, structural, and graph-based paradigms on ML-100K. For detailed metric descriptions, see Appendix~\ref{appendix: C}.}
\label{tab:stealthness_full_summary}
\resizebox{0.9\textwidth}{!}{
\begin{tabular}{lccccccccccc}
\toprule
\textbf{Method} & 
\multicolumn{2}{c}{\textbf{Mahalanobis Dist.}} & 
\multicolumn{2}{c}{\textbf{KDE Likelihood}} & 
\multicolumn{2}{c}{\textbf{OC-SVM Score}} & 
\multicolumn{2}{c}{\textbf{GAD}} & 
\textbf{RVC-Entropy} & 
\textbf{GSA Degree} & 
\textbf{IF Score} \\
\cmidrule(lr){2-3} \cmidrule(lr){4-5} \cmidrule(lr){6-7} \cmidrule(lr){8-9}
& Mean$\downarrow$ & Var & Mean$\uparrow$ & Var & Mean$\uparrow$ & Var & Mean$\downarrow$ & Var & Value$\uparrow$ & Value$\uparrow$ & Value$\uparrow$ \\
\midrule
Real Users     & 1.6284 & 0.3452 & 0.6881 & 0.1218 & -2.2533 & 198.4448 & 1.3580 & 0.2995 & N/A & 12.3471 & 0.0240 \\
CLeaR          & 1.5939 & 0.8466 & 0.5183 & 0.3185 & -2.6355 & 573.1659 & 1.1524 & 0.4062 & 0.5196 & 12.8889 & 0.0080 \\
AUSH           & 1.0273 & 0.0054 & 0.8739 & 0.0010 & 9.5084  & 2.1718   & 0.8503 & 0.0089 & 0.2416 & 15.5556 & 0.0509 \\
DLDA (w/o Dis) & 1.1033 & 0.0260 & 0.8371 & 0.0044 & 8.6781  & 8.5252   & 0.7986 & 0.0066 & 	0.2799 & 13.5556 & 0.0472 \\
\textbf{DLDA}  & 1.1332 & 0.0287 & 0.8247 & 0.0054 & 8.2378  & 10.5098  & 0.8133 & 0.0156 & 0.3911 & 15.2222 & 0.0468 \\
\bottomrule
\end{tabular}
}
\label{tab:stealthness_full_summary}
\end{table*}

To assess the stealthiness of DLDA-generated profiles, we conduct a comprehensive analysis from three complementary perspectives: (i) embedding space visualization, (ii) statistical alignment and diversity, and (iii) anomaly detection scores under multiple paradigms.

\textbf{Embedding Visualization.} Figure~\ref{fig:pca} shows a 3D PCA projection of the user embedding space. DLDA's generated users form a dispersed yet coherent cluster surrounding the real-user manifold. In comparison, AUSH and w/o Dis concentrate tightly near a centroid, while CLeaR drifts away from the genuine distribution. DLDA achieves a balanced spread between these extremes. Figure~\ref{fig:distance} quantifies this with pairwise centroid distances and intra-group variances. DLDA’s centroid distance to real users (0.2009) is lower than that of CLeaR (0.2834), indicating better semantic alignment, while its intra-group variance (0.0515) is higher than that of AUSH (0.0280), reflecting stronger behavioral diversity.

\textbf{Statistical Indicators.} Table~\ref{tab:stealthness_full_summary} reports six detection-related metrics. DLDA achieves low Mahalanobis distance (1.1332, var=0.0287) and high KDE likelihood (0.8247, var=0.0054), indicating close alignment with real-user distributions.
DLDA also shows moderate variance across metrics, avoiding the over-concentration of AUSH (KDE var=0.0010) and the instability of CLeaR (OC-SVM var=573.1659), thus balancing behavioral diversity and stealth.

\textbf{Graph and Structure-Aware Detection.} We further assess stealth against graph-based detectors that exploit topological patterns. DLDA achieves a moderate GNN-based contrastive score (0.8133), suggesting that fake users avoid forming overly unique or suspicious substructures. Its Isolation Forest score (0.0468) is comparable to that of w/o Dis and better than CLeaR (0.0080), demonstrating resistance to tree-based anomaly ranking. DLDA also achieves high average node degree (15.22), indicating realistic connectivity patterns, while preserving a moderate clustering coefficient. The elevated RVC-Entropy further indicates that DLDA avoids generating overly concentrated or homogeneous user clusters, benefiting from the dispersive regularization module.

Overall, DLDA consistently strikes a strong balance between semantic alignment and behavioral diversity. Unlike baselines that either collapse (AUSH) or over-disperse (CLeaR), DLDA generates fake users that blend smoothly into the real-user space, while evading a broad spectrum of anomaly detectors. These results validate the synergistic effect of dual guidance and dispersive regularization in enhancing stealth under structural and statistical scrutiny.

\subsection{Time \& Space Complexity Analysis (RQ5)}

We compare the computational cost of DLDA with two representative baselines, CLeaR and GSPAttack, focusing on time and space complexity. As shown in Table~\ref{tab:complexity_comparison}, DLDA offers a favorable trade-off between efficiency and scalability.

\textbf{Time Complexity.}
DLDA performs $T$ denoising steps and a lightweight Poisson-based projection without requiring gradient access or model retraining. Once trained, it generates fake users in parallel with a fixed, model-agnostic cost per sample, scaling only with diffusion steps and latent dimension. In contrast, CLeaR requires bi-level optimization and full-model access per injection, incurring high overhead as the user/item scale increases. GSPAttack is even more intensive, involving repeated GNN message passing, Gumbel-softmax sampling, and surrogate retraining when the victim model changes.

\textbf{Space Complexity.}
DLDA maintains compact latent representations and generates sparse rating vectors through lightweight projection, resulting in low memory usage. CLeaR stores dense fake profiles and intermediate gradients, causing significant memory overhead. GSPAttack stores a full GNN surrogate and produces dense rating vectors, making it less scalable to large datasets.

\begin{table}[t]
\centering
\caption{Time and space complexity comparison.}
\label{tab:complexity_comparison}
\resizebox{1.0\linewidth}{!}{
\begin{tabular}{lcll}
\toprule
\textbf{Attack Method} 
& \textbf{Time (s)} 
& \textbf{Time Complexity} 
& \textbf{Space Complexity} \\
\midrule
GSPAttack
& 14.03 
& $\mathcal{O}\big( E L (U+U_f+N) d^2 + E U_f N d^2 \big)$ 
& $\mathcal{O}\big( (U+U_f+N)d + U_f N \big)$ \\
CLeaR
& 121.03 
& $\mathcal{O}\big( E O (U+U_f) N d \big) + \mathcal{O}(U_f N \log K)$ 
& $\mathcal{O}\big( (U+U_f)d + U_f N \big)$ \\
\textbf{DLDA} 
& 15.03 
& $\mathcal{O}\big( U_f [\, T d^2 + N \log N ] \big)$ 
& $\mathcal{O}\big( U_f d + d^2 + N \big)$ \\
\bottomrule
\end{tabular}
}
\label{tab:complexity_comparison}
\end{table}

\section{Conclusion}

In this work, we examine the vulnerability of CF models to controllable and stealthy shilling attacks. We reveal that the difficulty in jointly achieving target promotion and behavioral realism stems from the lack of structure-aware, disentangled generation mechanisms. To address this, we propose DLDA, a latent diffusion-based attack framework that integrates conditional generation, dual cross-attention guidance, and dispersive regularization. Our method enables the synthesis of fake user profiles that are both target-effective and detection-evasive. Extensive experiments across multiple real-world datasets validate the effectiveness of DLDA in promoting target items without disrupting global recommendation quality. Our visual and quantitative analyses further confirm that DLDA generates diverse yet well-aligned profiles that blend into the genuine user space. In future work, we aim to explore robust defense strategies that counter such fine-grained controllable attacks in latent spaces.

\begin{acks}
The Australian Research Council supports this work under the streams of Future Fellowship (Grant No. FT210100624), the Discovery Project (Grant No. DP240101108), and the Linkage Project (Grant No. LP230200892).
\end{acks}

\bibliographystyle{ACM-Reference-Format}
\bibliography{mybib}

\appendix

\section{Information-Theoretic Intuition for Synergistic Design}
\label{appendix: B}
This section provides a detailed information-theoretic perspective to clarify how the combined conditional latent diffusion, dual cross-attention, and dispersive regularization modules work synergistically rather than being merely stacked.

\subsection{Overall Objective}

The goal is to generate fake user latent representations $\mathbf{z}_0^{\text{fake}}$ that are:
(i) aligned with the target user/item interaction patterns,
(ii) sufficiently diverse within the group,
(iii) indistinguishable from genuine user profiles.

Formally, we aim to:
\begin{equation}
\max I(\mathbf{z}_0^{\text{fake}};\, \mathbf{z}_u^c, \mathbf{z}_v^c)
\quad \text{s.t.} \quad
H(\mathbf{Z}^{\text{fake}}) \text{ is high.}
\end{equation}

\subsection{Conditional Latent Diffusion}

The conditional denoising process is trained to recover latent variables consistent with the target user/item conditions from noisy observations.
\begin{align}
\mathcal{L}_{\text{diff}} = 
\mathbb{E}_{\mathbf{z}_0, t, \boldsymbol{\epsilon}}
\Big[
\|\boldsymbol{\epsilon}_\theta(\sqrt{\bar{\alpha}_t} \mathbf{z}_0 
+ \sqrt{1 - \bar{\alpha}_t} \boldsymbol{\epsilon},\, t,\, 
\mathbf{z}_u^c,\, \mathbf{z}_v^c) - \boldsymbol{\epsilon}\|_2^2
\Big].
\end{align}
This shows that minimizing the denoising loss implicitly maximizes the information shared between generated fake users and the target user/item conditions, ensuring that the poisoned profiles exhibit behavioral alignment with the target preferences.

By minimizing this loss, the diffusion implicitly maximizes the mutual information between the generated representation and its conditioning variables:
\begin{equation}
\min \mathcal{L}_{\text{diff}} 
\quad \Longrightarrow \quad
\max I(\mathbf{z}_0^{\text{fake}};\, \mathbf{z}_u^c, \mathbf{z}_v^c).
\end{equation}

\subsection{Dual Cross-Attention}

At each diffusion step, dual cross-attention modules independently emphasize user and item conditions, ensuring that the generation remains guided by the target profiles.
\begin{align}
\mathbf{a}_u &= \text{CrossAttn}(\mathbf{q},\, \mathbf{z}_u^c), 
\quad
\mathbf{a}_v = \text{CrossAttn}(\mathbf{q},\, \mathbf{z}_v^c),\\
\mathbf{z}_t' &= \mathbf{z}_t + \mathbf{a}_u + \mathbf{a}_v.
\end{align}

By the data processing inequality:
\begin{equation}
I(\mathbf{z}_t';\, \mathbf{z}_u^c, \mathbf{z}_v^c) 
\geq I(\mathbf{z}_t;\, \mathbf{z}_u^c, \mathbf{z}_v^c),
\end{equation}
which means each residual attention step helps maintain or boost the information flow from the conditions.
This inequality guarantees that the dual cross-attention mechanism does not degrade the conditioning signal but reinforces it across time steps, improving controllability for target exposure.

\subsection{Dispersive Regularization}

To prevent local mode collapse and enforce intra-group diversity, we introduce the dispersive regularization:
\begin{equation}
\mathcal{L}_{\text{disp}} = 
\log \mathbb{E}_{i \neq j} 
\Big[
\exp(- \| \mathbf{m}_i - \mathbf{m}_j \|_2^2 / \tau)
\Big].
\end{equation}
Here, $\mathbf{m}_i$ and $\mathbf{m}_j$ denote the bottleneck representations (e.g., the latent feature vectors at the center of the U-Net denoiser) for different generated samples. The temperature $\tau$ controls the sharpness of repulsion: smaller $\tau$ enforces stronger penalization for close representations, thereby encouraging more separation. As a result, minimizing $\mathcal{L}_{\text{disp}}$ spreads out the fake user representations, indirectly maximizing the entropy of the generated group.

This acts as a soft lower bound to encourage high group entropy:
\begin{equation}
\min \mathcal{L}_{\text{disp}}
\; \Rightarrow \;
\max H(\mathbf{Z}^{\text{fake}}),
\quad
I(\mathbf{Z}^{\text{fake}};\,\mathbf{z}_u^c,\mathbf{z}_v^c) > 0.
\end{equation}

This effect can also be interpreted geometrically: maximizing the entropy of $\mathbf{Z}^{\text{fake}}$ is equivalent to encouraging a broader spread of representations across the latent space. Specifically, assuming $\mathbf{Z}^{\text{fake}}$ follows a Gaussian-like distribution, the entropy is closely tied to the log-determinant of the covariance matrix:
\begin{equation}
H(\mathbf{Z}^{\text{fake}}) \propto \log \det (\text{Cov}(\mathbf{Z}^{\text{fake}})).
\end{equation}

In this sense, minimizing $\mathcal{L}_{\text{disp}}$ indirectly promotes larger pairwise distances and higher covariance volume, which aligns to increase representation spread and avoid mode collapse.

Such covariance expansion is commonly used in representation learning to enhance sample diversity and coverage of the latent manifold, thus reducing the risk of overfitting and improving generalization.

\subsection{Formalized Lower Bound Linking to $\mathcal{L}_{\text{diff}}$ and $\mathcal{L}_{\text{disp}}$}

To make the information-theoretic intuition more concrete, we connect these objectives to the training losses with a provable lower bound.

\paragraph{Conditional Alignment.}
Using the InfoMax inequality \cite{barber2004algorithm}:
\begin{align}
I(\mathbf{z}_0^{\text{fake}};\, \mathbf{z}_u^c, \mathbf{z}_v^c) 
&\geq H(\mathbf{z}_0^{\text{fake}}) 
- H(\mathbf{z}_0^{\text{fake}} |\, \mathbf{z}_u^c, \mathbf{z}_v^c), \\
\mathcal{L}_{\text{diff}}  &\approx
H(\mathbf{z}_0^{\text{fake}} |\, \mathbf{z}_u^c, \mathbf{z}_v^c)
 .
\end{align}

This approximation is inspired by the variational inference view of denoising diffusion models \cite{sohl2015deep, ho2020denoising}, where the training objective minimizes the expected reconstruction error of the underlying noise. Since noise is sampled independently of conditions, the residual error effectively measures the uncertainty—or conditional entropy—of the generated representation given the conditioning inputs. Thus, $\mathcal{L}_{\text{diff}}$ serves as a tractable surrogate for $H(\mathbf{z}_0^{\text{fake}} |\, \mathbf{z}_u^c, \mathbf{z}_v^c)$.

\paragraph{Diversity via Dispersive Loss.}
By Jensen's inequality:
\begin{align}
H(\mathbf{Z}^{\text{fake}})
&\geq - \log\, 
\mathbb{E}_{i \neq j}
\Big[
\exp(- \| \mathbf{m}_i - \mathbf{m}_j \|_2^2 / \tau)
\Big] \notag\\
&= - \mathcal{L}_{\text{disp}}.
\end{align}

\paragraph{Combined Lower Bound.}
Combining these gives:
\begin{align}
I(\mathbf{z}_0^{\text{fake}};\, \mathbf{z}_u^c, \mathbf{z}_v^c) 
+ H(\mathbf{Z}^{\text{fake}})
&\;\gtrsim\; - \mathcal{L}_{\text{diff}} - \mathcal{L}_{\text{disp}}.
\end{align}
This implies that our loss minimization objective jointly optimizes target preference alignment and behavioral diversity in an information-theoretic sense.

\subsection{Synergistic Summary}
The three modules complement each other:
\begin{itemize}[leftmargin=*]
\item \textbf{Conditional Diffusion} aligns generated samples with the specified target user/item conditions.
\item \textbf{Dual Cross-Attention} dynamically injects user and item guidance at each timestep to enhance conditional fidelity.
\item \textbf{Dispersive Loss} promotes intra-group diversity and alleviates local mode collapse.
\item[$\rightarrow$] Together, these components produce controllable and stealthy fake user profiles that are both realistic and target-oriented.
\end{itemize}

In summary, our design maximizes target-oriented mutual information and preserves diversity in the latent space, ensuring that the generated fake user profiles blend naturally into the genuine user manifold.

\section{Stealthiness Metric Descriptions}
\label{appendix: C}

To comprehensively assess the stealthiness of generated users, we consider evaluation metrics spanning multiple perspectives:

\begin{itemize}[leftmargin=*]
    \item \textbf{Mahalanobis Distance ($\downarrow$)}: Measures the statistical deviation of fake users from the distribution of real users in the embedding space, using the inverse covariance matrix for normalization.
    
    \item \textbf{KDE Likelihood ($\uparrow$)}: Estimates the likelihood of a fake user under the kernel density distribution fitted on real users, indicating conformity to data density.
    
    \item \textbf{OC-SVM Score ($\uparrow$)}: Uses one-class SVM trained on real users to assign anomaly scores to generated users.
    
    \item \textbf{GAD ($\downarrow$)}: Detection confidence score from a GNN-based contrastive anomaly detector. This method leverages user-item interaction graphs and contrastive learning to distinguish fake users from real ones. 
    
    \item \textbf{RVC-Entropy ($\uparrow$)}: Quantifies the classification uncertainty of fake users under a KNN-based binary classifier trained to distinguish real from fake users. A higher RVC-Entropy reflects increased classification confusion between real and fake users, indicating that the synthetic profiles exhibit stronger stealth and realism.
    
    \item \textbf{GSA Degree ($\uparrow$)}: Refers to the average node degree of fake users in a $k$NN graph built from user embeddings, reflecting structural integration.
    
    \item \textbf{IF Score ($\uparrow$)}: Measures the Local Influence Function (LIF)-based normality score, estimating how typical a fake user appears when modeled as a data point in an influence framework.
\end{itemize}

Higher values indicate better stealthiness for all metrics except those marked with ($\downarrow$), where lower is better.

\end{document}